\documentclass{article}

\usepackage[nonatbib, final]{nips_2017}

\usepackage{times}
\usepackage{epsfig}
\usepackage{graphicx}
\usepackage{amsmath}
\usepackage{amssymb}
\usepackage{dsfont}
\usepackage{siunitx}
\usepackage{subfigure}
\usepackage{multirow}
\usepackage{caption}
\usepackage{booktabs}       
\usepackage{standalone}
\usepackage{xspace}
\usepackage{color}
\usepackage{comment}

\graphicspath{{./images_lowres/}}

\def\argmax{\mathop{\rm argmax}}

\def\1{\mathds{1}}

 \makeatletter
 \DeclareRobustCommand\onedot{\futurelet\@let@token\@onedot}
 \def\@onedot{\ifx\@let@token.\else.\null\fi\xspace}
  
 \def\ie{i.e\onedot}

 \makeatother

\newcommand{\myparagraph}[1]{\vspace{3pt}\noindent{\bf #1}}

\usepackage[pagebackref=true,breaklinks=true,letterpaper=true,colorlinks,bookmarks=false]{hyperref}

\begin{document}

\title{Attentive Explanations: Justifying Decisions and Pointing to the Evidence (Extended Abstract)}

 \newcommand{\authSpace}{&}
 \author{
 Dong Huk Park$^{1}$, Lisa Anne Hendricks$^{1}$, Zeynep Akata$^{2,3}$, Anna Rohrbach$^{1,3}$, \\
 \textbf{Bernt Schiele$^{3}$, Trevor Darrell$^{1}$, and Marcus Rohrbach$^{4}$} \vspace{2mm} \\
  $^{1}$EECS, UC Berkeley,  $^2$University of Amsterdam,
 $^{3}$MPI for Informatics, $^{4}$Facebook AI Research \\
}

\maketitle

\begin{abstract}
Deep models are the defacto standard in visual decision problems due to their impressive performance on a wide array of visual tasks.
On the other hand, their opaqueness has led to a surge of interest in explainable systems. In this work, we emphasize the importance of model explanation in various forms such as visual pointing and textual justification. The lack of data with justification annotations is one of the bottlenecks of generating multimodal explanations.
Thus, we propose two large-scale datasets with annotations that visually and textually justify a classification decision for various activities, i.e. ACT-X, and for question answering, i.e. VQA-X.
We also introduce a multimodal methodology for generating visual and textual explanations simultaneously. 
We quantitatively show that training with the textual explanations not only yields better textual justification models, but also models that better localize the evidence that support their decision.
\end{abstract}

\section{Introduction}
\label{sec:intro}

Explaining decisions is an integral part of human communication, understanding, and learning.
Therefore, we aim to build models that explain their decisions, something which comes naturally to humans. 
Explanations can take many forms.
For example, humans can explain their decisions with natural language, or by pointing to visual evidence.

We show here that deep models can demonstrate similar competence, and develop a novel multi-modal model which textually justifies decisions and visually grounds evidence simultaneously. 

To measure the quality of the generated explanations, compare with different methods, and understand when methods will generalize, it is important to have access to ground truth human annotations. Unfortunately, there is a dearth of datasets which include examples of how humans justify specific decisions. We thus propose and collect explanation datasets for two challenging vision problems: activity recognition and visual question answering (VQA).

We confirm whether the model is actually attending to the discussed items when generating the textual justification (as opposed to just memorizing justification text) by comparing it to our visual pointing annotations. We also determine whether the model attends to the same regions when making a decision as it does when explaining its decision.


\section{Multimodal Explanation Datasets}
\label{sec:datasets}

\begin{figure}[t]
\center
  \includegraphics[width=0.48\linewidth]{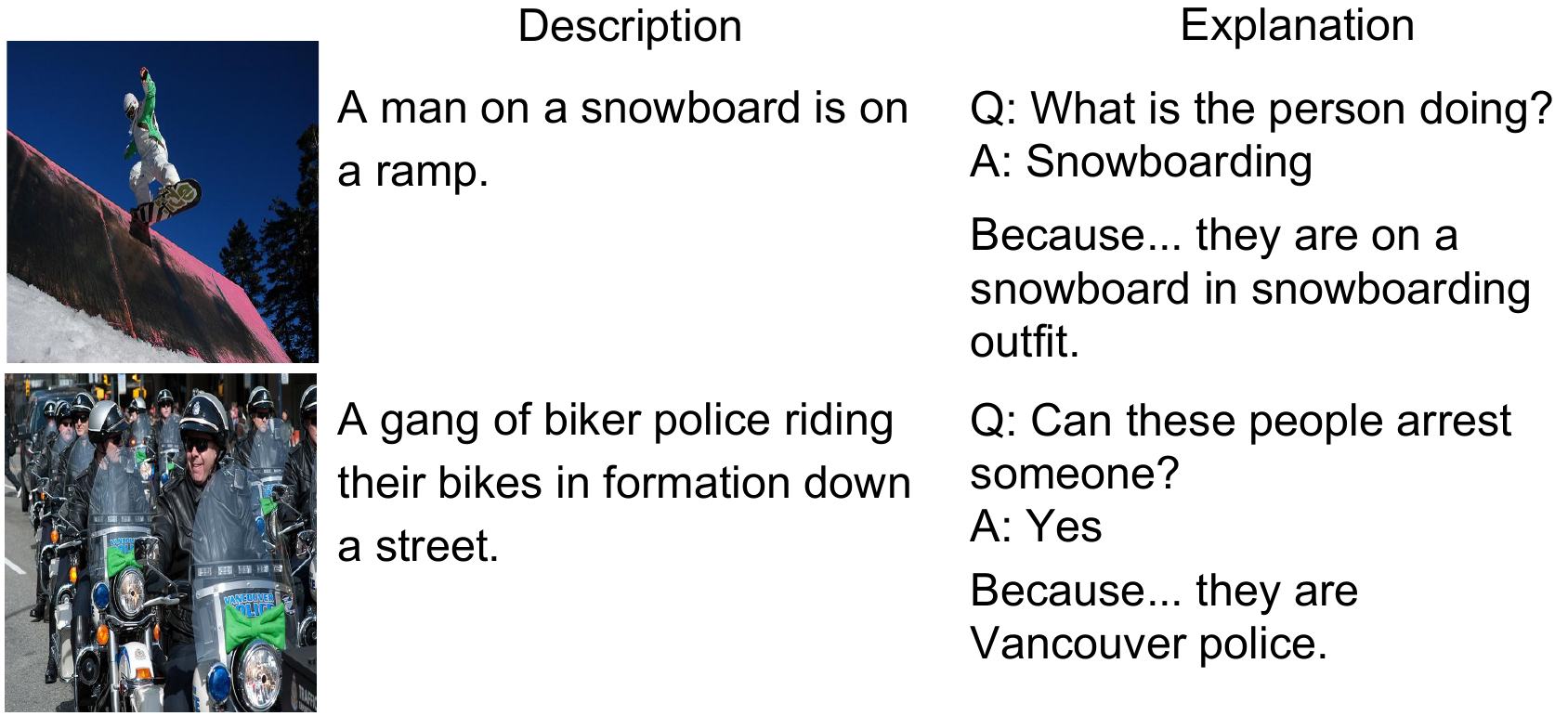}\hfill      \includegraphics[width=0.48\linewidth]{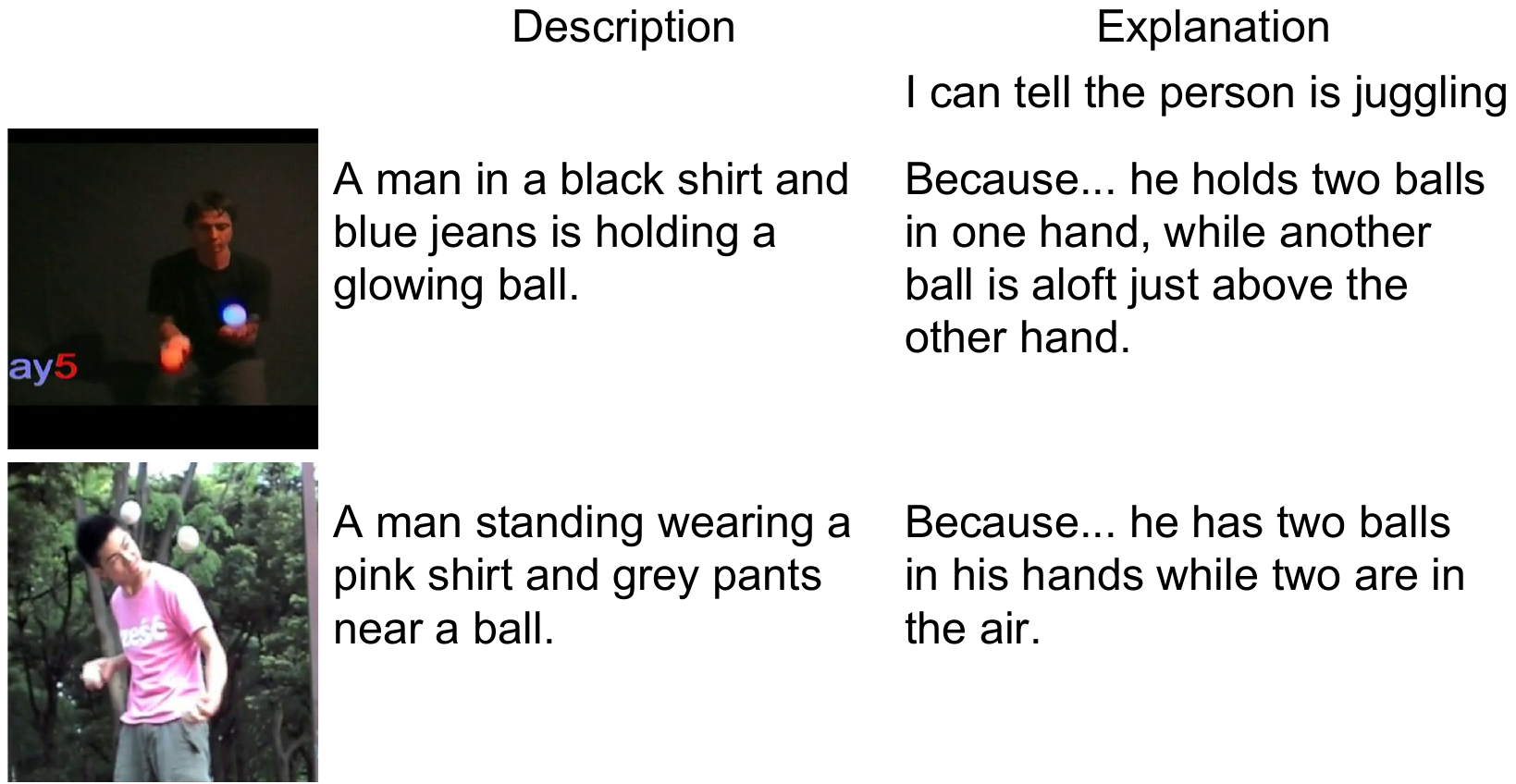}
\caption{(Left) Our VQA-X explanations focus on the visual evidence that pertains to the question and answer instead of generally describing objects in the scene. (Right) Our ACT-X explanations are task specific whereas image descriptions are more generic.} 
\label{fig:VQA_examples}
\end{figure}

\myparagraph{VQA Explanation Dataset (VQA-X).} The VQA dataset~\cite{antol2015vqa} contains open-ended questions about images which require understanding vision, natural language, and commonsense knowledge to answer.
We collected $1$ explanation per data point for a subset of the training set and $5$ explanations per data point for a subset of the validation and test sets, summing up to $30$k justification sentences. The annotators were asked to provide a proper sentence or clause that would come after the proposition ``because'' as explanations to the provided image, question, and answer triplet. Examples for both descriptions, \ie from MSCOCO dataset\cite{coco2014}, and our explanations are presented in \autoref{fig:VQA_examples}.

\myparagraph{Activity Explanation Dataset (ACT-X).} The MPII Human Pose (MHP) dataset~\cite{APGS14} contains images extracted from videos downloaded from Youtube.
For each image we collected $3$ explanations, totaling $54$k sentences. During data annotation, we asked the annotators to complete the sentence ``I can tell the person is doing X because..'',  where X is the ground truth activity label, see \autoref{fig:VQA_examples}.

\begin{figure*}
\centering
	\subfigure[VQA-X]{\label{fig:vqa_seg_gt}\includegraphics[height=20mm]{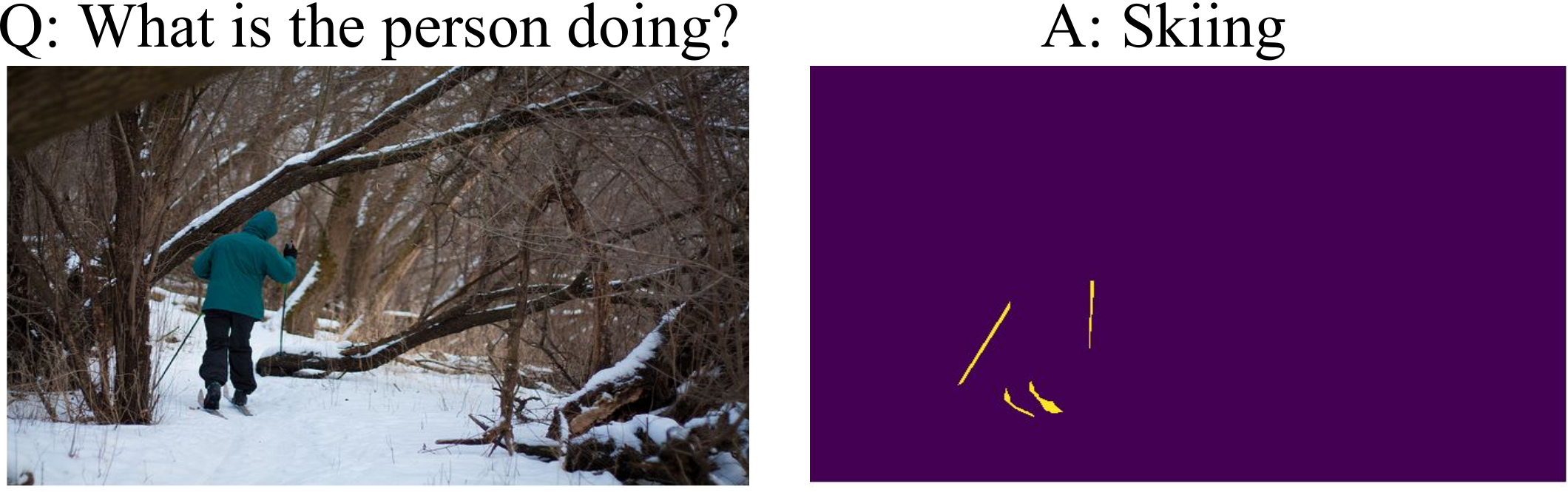}}
\subfigure[ACT-X]{\label{fig:act_seg_gt}\includegraphics[height=20mm]{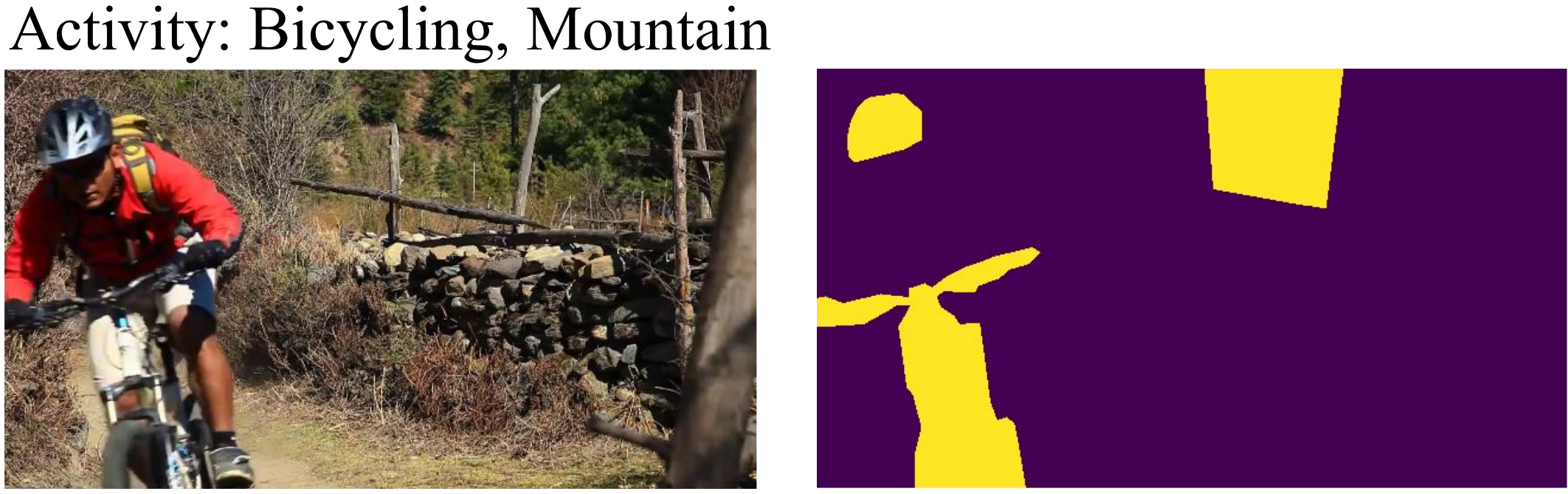}}
\vspace{-0.3cm}
\caption{Example human visual explanations collected on: (left)  VQA-X dataset, (right) ACT-X dataset. The visual evidence that justifies the answer is segmented in yellow.}
\label{fig:GroundTruthMap}
\end{figure*}

\myparagraph{Visual Pointing.}
In addition to textual justification, we collect visual explanations from humans for both VQA-X and ACT-X datasets. 
Annotators are provided with an image and an answer (question and answer pair for VQA-X, class label for ACT-X). They are asked to segment objects and/or regions that most prominently justify the answer.
Some examples can be seen in \autoref{fig:GroundTruthMap}.

\begin{figure*}[t]
  \includegraphics[width=\linewidth]{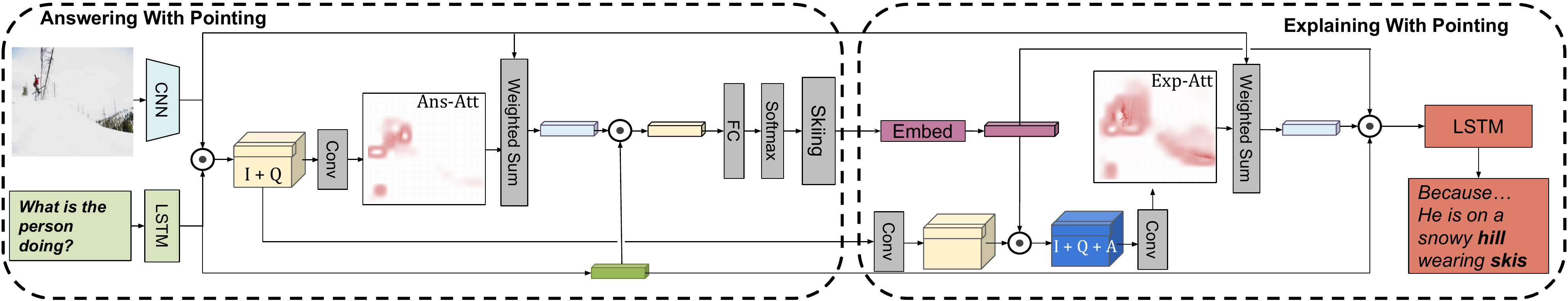}
\caption{Our Pointing and Justification (PJ-X)
architecture generates a multi-modal explanation which includes a textual justification (``He is on a snowy hill wearing skis'') and points to the visual evidence.  Our model consists of two ``pointing'' mechanisms: answering with pointing (left) and explaining with pointing (right).}
\vspace{-0.3cm}
\label{fig:attentive_explanation}
\end{figure*}

\section{Pointing and Justification Model (PJ-X)}
\label{sec:method}

The goal of our work is to justify why a decision was made with natural language, and point to the evidence for both the decision and the textual justification provided by the model.
We deliberately design our Pointing and Justification Model (PJ-X) to allow training these two tasks as well as the decision process jointly.
Specifically we want to rely on natural language justifications and the classification labels as the only supervision. 
We design a model which learns ``to point'' in a latent way. For the pointing we rely on an attention mechanism \cite{bahdanau2014neural} which allows the model to focus on a spatial subset of the visual representation. Our model uses two different attentions: one for making predictions and another for generating textual explanations. 
%
%
%
We first predict the answer given an image and a question. Then given the answer, question, and image, we generate the textual justification. In both cases we include a latent attention mechanism which allows to introspect where the model is looking. 
An overview of our double attention model is presented in \autoref{fig:attentive_explanation} and the detailed formulation is given in \cite{park2016arxiv}.

\section{Experiments}
\label{sec:exp}

\myparagraph{Experimental Setup.}
\label{sec:experimentalsetup}
For visual question answering, our model is pre-trained on the VQA training set \cite{antol2015vqa} to achieve state-of-the-art performance, but we either freeze or finetune the weights of the prediction model when training on explanations as the VQA-X dataset is significantly smaller than the original VQA training set. We refer the finetuned model as `Finetuned' throughout the paper. For activity recognition, prediction and explanation components of the model are trained jointly.

{
\setlength{\tabcolsep}{3pt}
\renewcommand{\arraystretch}{1.2}
\begin{table*}[tb]
\begin{center}
\footnotesize
\begin{tabular}{l l l l|rrrr r |rrrr r}
\toprule
&Train-&Att.&Answer& \multicolumn{5}{c|}{VQA-X} & \multicolumn{5}{c}{ACT-X} \\
&ing &  for & Condi-& \multicolumn{4}{c}{Automatic eval.} & \multicolumn{1}{c|}{Human} &\multicolumn{4}{c}{Automatic eval} & \multicolumn{1}{c}{Human}\\
Approach &  Data & Expl. &  tioning & \multicolumn{1}{c}{B} & \multicolumn{1}{c}{M} & \multicolumn{1}{c}{C} &\multicolumn{1}{c}{S} &\multicolumn{1}{c|}{eval.} & \multicolumn{1}{c}{B} & \multicolumn{1}{c}{M} & \multicolumn{1}{c}{C} &\multicolumn{1}{c}{S} &\multicolumn{1}{c}{eval}  \\
\midrule
\cite{hendricks16eccv} & Desc. & No & Yes & 	\multicolumn{1}{c}{--}&\multicolumn{1}{c}{--}&\multicolumn{1}{c}{--}&\multicolumn{1}{c}{--}&\multicolumn{1}{c|}{--}&12.9 & 15.9& 12.4 & 12.0 &7.6  \\ 
Ours on Descript. & Desc. & Yes & Yes & 8.1 & 14.3  & 34.3 & 11.2 &24.0 & 6.9  & 12.9 & 20.3 & 7.3&18.0  \\ 
\midrule
Captioning Model & Expl. & Yes & No &17.1 &16.0  &43.6 &7.3 &19.2 &20.7 &18.8  &40.7 &11.3&20.4 \\ 
Ours w/o Exp-Att. & Expl. & No & Yes &25.1 &20.5  &74.2 &11.6 &34.4& 16.9 &17.0 &33.3 &10.6 &17.6  \\ 
Ours & Expl. & Yes & Yes &25.3 &20.9   &72.1 &12.1 & 33.6 & 24.5 &21.5  &58.7 &16.0 & 26.4  \\ 
Ours (Finetuned) & Expl. & Yes & Yes &27.1 &20.9   &77.2 &11.8  & -- & -- & -- & -- & -- & --  \\
\bottomrule
\end{tabular}
\vspace{-2mm}
\caption{Evaluation of Textual Justifications. Evaluated automatic metrics:  BLEU-4 (B), METEOR (M), CIDEr (C), SPICE (S). Human evaluation: 250 random images 3 judges rate whether a generated explanation is better than, worse than, or equivalent to a ground truth explanation. We report the \% of generated explanations which are equivalent to or better than ground truth human explanations, when at least 2 out of 3 human judges agree.}
\vspace{-4mm}
\label{tbl:generation_automatic}
\end{center}
\end{table*}
}

\myparagraph{Textual Justification.}
\label{sec:res:textual}
We ablate our model and compare with related approaches on our VQA-X and ACT-X datasets based on automatic and human evaluation for the generated explanations.

We re-implemented the state-of-the-art captioning model~\cite{donahue16pami} with an integrated attention mechanism which we refer to as ``Captioning Model''. This model only uses images and does not use class labels (i.e. the answer in VQA-X and the activity label in ACT-X) when generating textual justifications.
We also compare with~\cite{hendricks16eccv} using publicly available code.
Note that \cite{hendricks16eccv} is trained with image descriptions and justifications are generated conditioned on both the image and the class predictions.
``Ours on Descriptions'' is another ablation in which we train the PJ-X model on descriptions instead of explanations.
``Ours w/o Exp-Attention'' is similar to~\cite{hendricks16eccv} in the sense that there is no attention mechanism for generating explanations, however, it does not use the discriminative loss and is trained on explanations instead of descriptions.

Our PJ-X model performs well when compared to the state-of-the-art on both automatic evaluation metrics and human evaluations (Table \ref{tbl:generation_automatic}). ``Ours'' model significantly improves ``Ours with description'' model by a large margin on both datasets.
Additionally, our model outperforms \cite{hendricks16eccv} which learns to generate explanations given only description training data.  These results confirm that our new datasets with ground truth explanations are important for textual justification generation.
Comparing ``Ours'' to ``Captioning Model'' shows that conditioning explanations on a model decision is important (human evaluation score increases from 20.4 to 26.4 on ACT-X and 19.2 to 33.6 on VQA-X).
Thus it is important for our model to have access to questions and answers to accurately generate the explanation.
Finally, including attention allows us to build a multi-modal explanation model.
On the ACT-X dataset, it is clear that including attention (compare ``Ours w/o Exp-Attention'' to ``Ours'') greatly improves textual justifications.
On the VQA-X dataset, ``Ours w/o Attention'' and ``Ours'' are comparable, however,
the latter also produces a multi-modal explanation that offers us an added insight about a model's decision.

\begin{table}[tb]
\footnotesize
\centering
\begin{tabular}{l|rr|rrr}
\toprule
& \multicolumn{2}{c|}{Earth Mover's distance } & \multicolumn{3}{c}{ Rank Correlation }  \\
& \multicolumn{2}{c|}{(lower is better)} & \multicolumn{3}{c}{(higher is better) }  \\

& VQA-X & ACT-X & VQA-X & ACT-X & VQA-HAT \\
\midrule

Random Point & 9.21 & 9.36 & -0.0010 & +0.0003 & -0.0001 \\
Uniform & 5.56 &  4.81  & -0.0002 & -0.0007 & -0.0007 \\
HieCoAtt-Q~\cite{DBLP:journals/corr/DasAZPB16} & -- & -- & -- & -- & 0.2640 \\
Ours (ans-att) & 4.24 & 6.44  & +0.2280 & +0.0387 & +0.1366 \\
Ours (exp-att) & 4.31 & \textbf{3.8}  & +0.3132 & \textbf{+0.3744} & +0.3988 \\
Finetuned (ans-att) & \textbf{4.24} & --  & +0.2290 & -- & +0.2809 \\
Finetuned (exp-att) & \textbf{4.25} & -- & \textbf{+0.3152} & -- & \textbf{+0.5041} \\

\bottomrule
\end{tabular}
\caption{Evaluation of visual pointing. Ours (ans-att) denotes the attention map used to predict the answer whereas Ours (exp-att) denotes the attention map used to generate explanations.
}
\vspace{-0.3cm}
\label{tbl:attention_map_eval_wo}
\label{tbl:attention_map_eval_rank_corr}
\end{table}

\myparagraph{Visual Pointing.}
\label{sec:res:pointing}
We compare our generated attention maps to the following baselines and report quantitative results with corresponding analysis.

\textit{Random Point} randomly attends to a single point in a $14\times 14$ grid. \textit{Uniform Map} generates attention map that is uniformly distributed over the $14\times 14$ grid.
We denote the attention map used to predict the answer as {\em ans-att}, whereas {\em exp-att} denotes the attention map used to generate explanations.

We evaluate attention maps using the Earth Mover's Distance (lower is better) and rank correlation (higher is better) on VQA-X and ACT-X datasets in~\autoref{tbl:attention_map_eval_wo}.
We observe that our exp-att outperforms baselines, indicating that exp-att aligns well with human annotated explanations.
For ACT-X, our exp-att also outperforms all the baselines. The exp-att significantly outperforms the ans-att, indicating that the regions the model attends to when generating an explanation agree more with regions humans point to when justifying a decision. This suggests that whereas ans-att attention maps can be helpful for understanding a model and debugging, they are not necessarily the best option when providing visual evidence which agrees with human justifications

\section{Conclusion}
\label{sec:conc}

As a step towards explainable AI models, we introduced two novel explanation datasets collected through crowd sourcing for visual question answering and activity recognition, \ie VQA-X and ACT-X. We also proposed a multimodal explanation model that is capable of providing natural language justifications of decisions and pointing to the evidence. We quantitatively demonstrated that both attention and supervision from the reference justifications help achieve high quality textual and visual explanations.

\bibliographystyle{ieee}
\bibliography{biblioShort,biblio,egbib}

\end{document}


\title{Attentive Explanations: Justifying Decisions and Pointing to the Evidence \\ (Supplementary Material)}

\author{First Author\\
Institution1\\
Institution1 address\\
{\tt\small firstauthor@i1.org}
\and
Second Author\\
Institution2\\
First line of institution2 address\\
{\tt\small secondauthor@i2.org}
}

\newcommand{\toAdd}[1]{\textcolor{red}{\textbf{Todo #1}}}
\maketitle
\thispagestyle{empty}

\begin{figure*}
\centering
\subfigure[VQA-X]{\label{fig:vqa_seg_gt}\includegraphics[width=80mm]{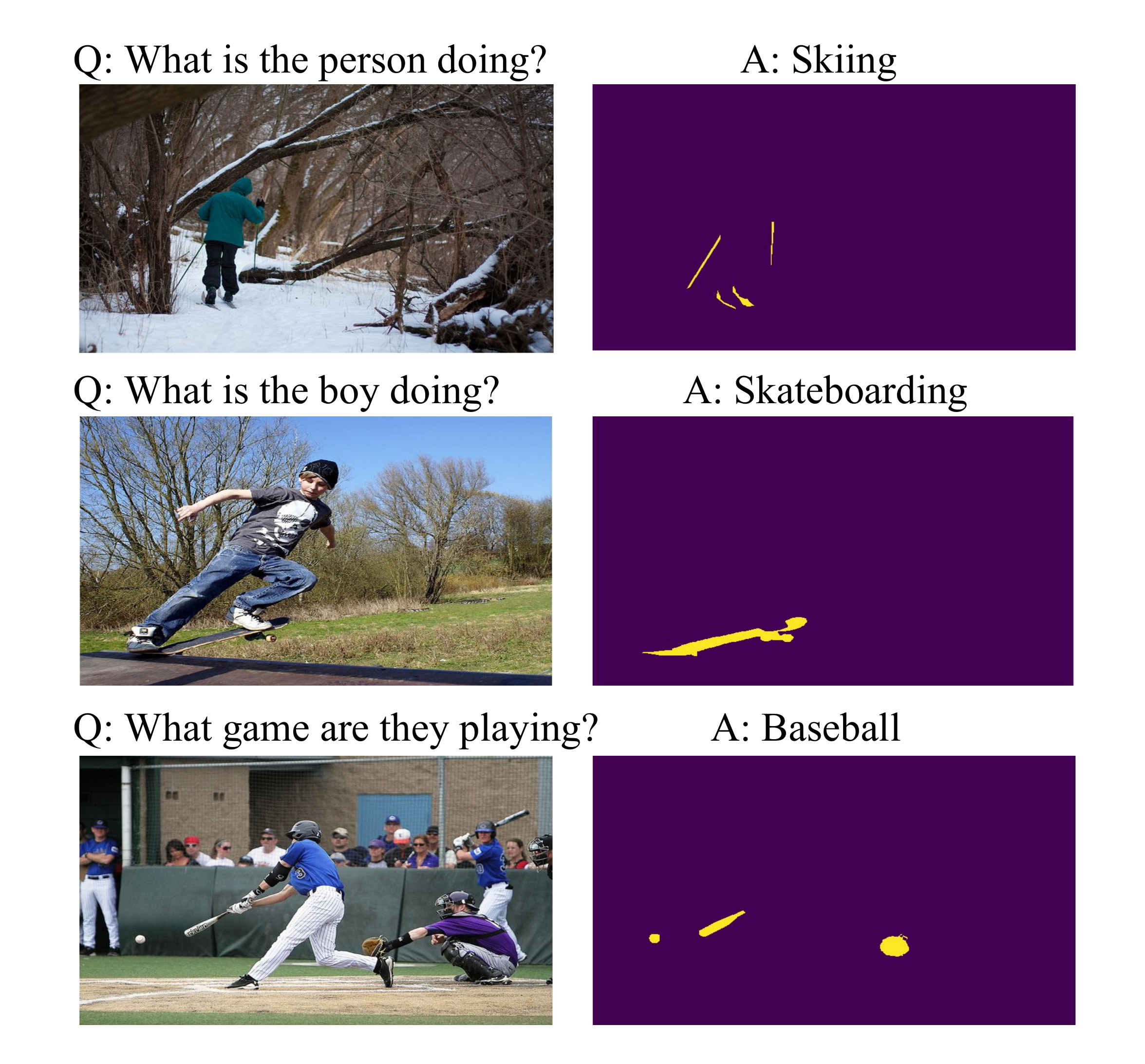}}
\subfigure[ACT-X]{\label{fig:act_seg_gt}\includegraphics[width=80mm]{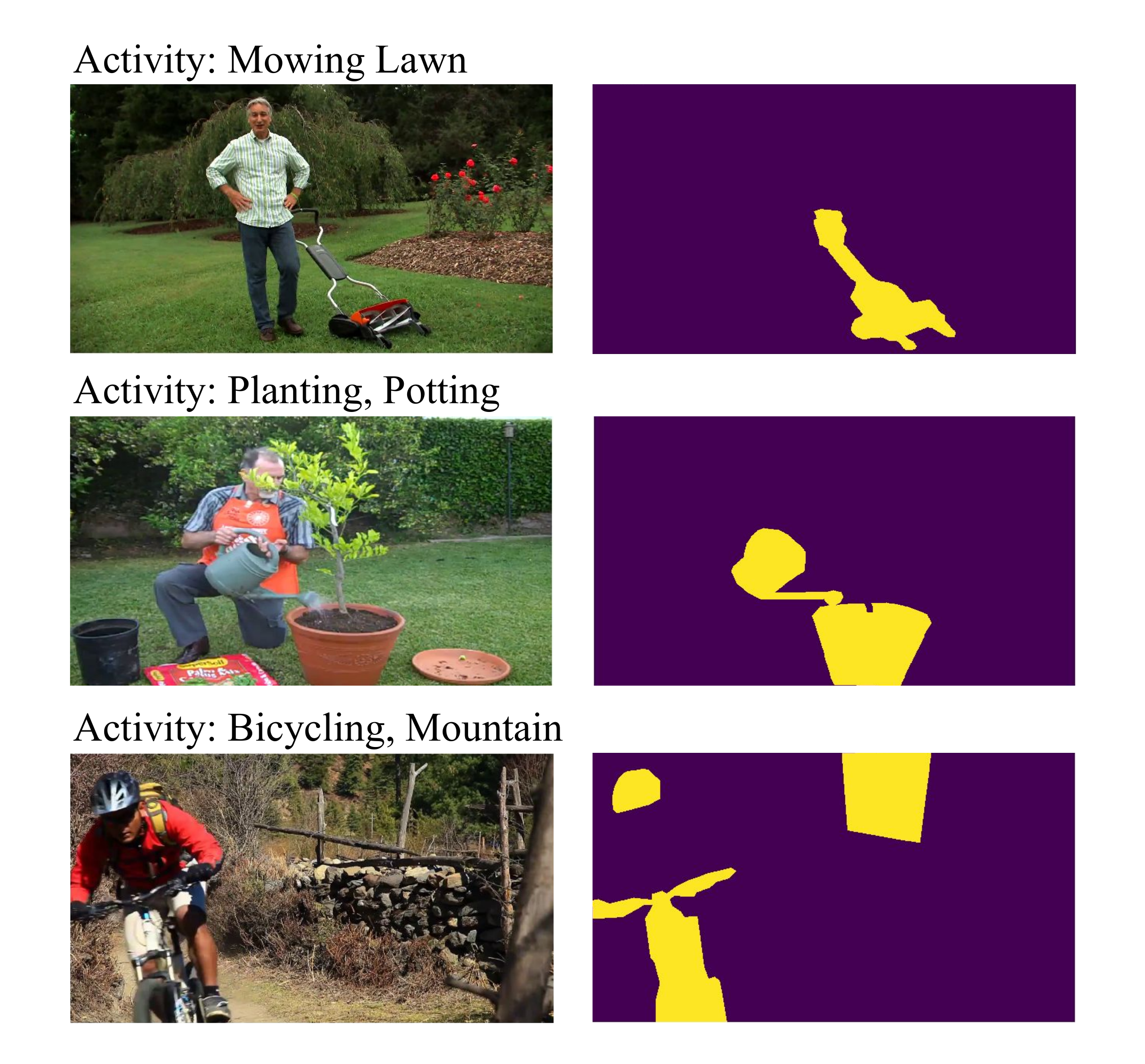}}
\caption{Human-Annotated Attention Maps. Figure on the left: Example annotations collected on VQA-X dataset for the pointing task. Figure on the right: Example annotations collected on ACT-X dataset for the pointing task. In both cases, the visual evidence that justifies the answer is segmented in yellow.}
\label{fig:VQAqualitative}
\end{figure*}

In this supplemental material, we provide details about our human-annotated attention maps used for quantitative evaluation of the pointing task (Section \ref{sec:human_att_maps}, Figure \ref{fig:VQAqualitative}), as well as additional qualitative results (Section \ref{supp:qualitative},  Figures \ref{fig:VQAqualitative-sameQA},  \ref{fig:VQAqualitative-sameimage}, \ref{fig:ACTqualitative:similar:activity}, \ref{fig:VQAqualitative:correct_incorrect}, and  \ref{fig:ACTqualitative:correct_incorrect}).

\section{Human-Annotated Attention Maps}
\label{sec:human_att_maps}

In order to evaluate if the
attention of our model corresponds to where humans think
the evidence for the answer is, we collect
attention maps from humans for both VQA-X and ACT-X
datasets (Section 4 of the main paper). Human-annotated attention maps are collected via
Amazon Mechanical Turk where we use the segmentation UI interface from the OpenSurfaces Project \cite{bell13opensurfaces}. Annotators are provided with an image and an answer (question and answer pair for VQA-X, class label for ACT-X). They are asked to segment objects and/or regions that most prominently justify the answer. For each dataset we randomly sample 500 images from the test split, and for each image we collect 3 attention maps. The collected annotations are used in the main paper for computing the Earth Mover's Distance to evaluate attention maps of our model against several baselines. 

\section{Qualitative Results}
\label{supp:qualitative}

Figures \ref{fig:VQAqualitative-sameQA} 
and \ref{fig:VQAqualitative-sameimage} 
demonstrate that both images and the question/answer pair are needed for good explanations.
Figure \ref{fig:VQAqualitative-sameQA} shows explanations for different images, but with the same question/answer pair.
Importantly, explanation text and visualizations change to reflect image content. For instance, for the question ''Where is this picture taken?'' our model explains the answer ''Airport'' by pointing and discussing planes and trucks in the first image while pointing and discussing baggage carousel in the second image.
Figure  \ref{fig:VQAqualitative-sameimage} shows that when different questions are asked about the same images, explanations provide information which are specific to the questions. For example, for the question ''Is it sunny?'' our model explains the answer ''Yes'' by mentioning the sun and its reflection and pointing to the sky and the water, whereas for the question ''What is the person doing?'' it points more directly to the surfer and mentions that the person is on a surfboard. 

Figure \ref{fig:ACTqualitative:similar:activity} shows that explanations on the ACT-X dataset discuss small details important for differentiating between similar classes.  For example, when explaining kayaking and windsurfing, it is important to mention the correct sporting equipment such as ''kayak'' and ''sail'' instead of image context. On the other hand, when distinguishing bicycling (BMX) and bicycling (racing and road), it is important to discuss the image context such as ''doing a trick on a low wall'' and ''riding a bicycle down the road.''

\begin{figure*}[h]
\begin{center}
  \includegraphics[width=\linewidth]{supp_sameqa_diffimg}
\end{center}
\caption{VQA-X results with the same question/answer pair. We select results with the same question and answer pair with two different images and show that although the QA pairs are the same, for different images our model generates different explanations (Answers are correctly predicted). VQA-ATT denotes attention maps used for predicting answers and EXP-ATT denotes attention maps used for generating the corresponding justifications.}
\label{fig:VQAqualitative-sameQA}
\end{figure*}

\newpage



\begin{figure*}[h]
\begin{center}
  \includegraphics[width=\linewidth]{supp_sameimg_diffqa}
\end{center}
\caption{VQA-X results with same image and different questions. We select results with the same image and different Q/A pairs and show that although the images are the same, our model is able to answer the questions differently and generate a different explanation accordingly (Answers are correctly predicted). VQA-ATT denotes attention maps used for predicting answers and EXP-ATT denotes attention maps used for generating the corresponding justifications.}
\label{fig:VQAqualitative-sameimage}
\end{figure*}

\newpage

Figures \ref{fig:VQAqualitative:correct_incorrect} and \ref{fig:ACTqualitative:correct_incorrect} compare explanations when the answers or action labels are correctly and incorrectly predicted. In addition to providing an intuition about why predictions are correct, our explanations frequently justify why the model makes incorrect predictions. For example, when incorrectly predicting whether
one should stop or go (Figure 5, lower-right example), the model outputs ''Because the light is green'' suggesting that the model has mistaken a red light for a green light, and furthermore, that green lights mean ''go''.

Figure \ref{fig:ACTqualitative:correct_incorrect} shows similar trends on the ACT-X dataset. For example, when incorrectly predicting the activity power yoga for an image depicting manual labor, the explanation ''Because he is sitting on a yoga mat and holding a yoga pose” suggests that the rug may have been misclassified as a yoga mat. We reiterate that our model justifies predictions and does not fully explain the inner-workings of deep architectures. However, these justifications demonstrate that our model can output intuitive explanations which could help those unfamiliar with deep architectures make sense of model predictions.

\begin{figure*}[h]
\begin{center}
 \includegraphics[width=\linewidth]{supp_sameact_diffimg}
\end{center}
\caption{ACT-X results with similar activities. Figure on the left: We show results with fine-grained activities all related to windsurfing, kayaking, canoeing and observe that both the fine-grained activities are correctly predicted and the explanations match the activity and the image. Figure on the right: We show results with fine-grained activities all related to bicycling and observe that both the fine-grained activities are correctly predicted and the explanations match the activity and the image. ACT-ATT denotes attention maps used for predicting answers and EXP-ATT denotes attention maps used for generating the corresponding justifications.}
\label{fig:ACTqualitative:similar:activity}
\end{figure*}

\begin{figure*}[h]
\begin{center}
 \includegraphics[width=\linewidth]{vqa_right_wrong}
\end{center}
\caption{VQA-X results. GT denotes ground-truth answer while P indicates actual prediction made by the model. Figure on the left: We show various qualitative results with correctly predicted answer and observe that the explanation justifies the answer accordingly. Figure on the right: We show results with incorrectly predicted answer and observe that although the answer is incorrect, our model can provide visual and textual explanations on why the model might be failing in those cases.}
\label{fig:VQAqualitative:correct_incorrect}
\end{figure*}

\begin{figure*}[h]
\begin{center}
 \includegraphics[width=\linewidth]{activity_right_wrong}
\end{center}
\caption{ACT-X results. GT denotes ground-truth answer while P indicates actual prediction made by the model. Figure on the left: We show various qualitative results with correctly predicted answer and observe that the explanation justifies the answer accordingly. Figure on the right: We show results with incorrectly predicted answer and observe that although the answer is incorrect, our model can provide visual and textual explanations on why the model might be failing in those cases.}
\label{fig:ACTqualitative:correct_incorrect}
\end{figure*}

\bibliographystyle{ieee}
\bibliography{biblioLong,biblio,egbib}